\documentclass[letterpaper, 10 pt, conference]{ieeeconf}  
\IEEEoverridecommandlockouts                              
\usepackage{times}
\usepackage{amsmath,amsfonts}
\usepackage{algorithmicx}
\usepackage{algorithm}
\usepackage{array}
\usepackage{textcomp}
\usepackage{stfloats}
\usepackage{url}
\usepackage{verbatim}
\usepackage{todonotes}
\usepackage{subcaption}
\usepackage{graphicx} 
\usepackage{xcolor}
\usepackage{threeparttable}
\usepackage{multirow}
\usepackage{amsthm}
\usepackage{mathrsfs}
\usepackage{amssymb}
\usepackage{algpseudocode}
\usepackage{url}
\usepackage{hyperref}
\usepackage{makecell} 
\usepackage{amssymb}  
\usepackage{algorithmicx}
\usepackage{algpseudocode}
\usepackage{multicol}
\usepackage{color,soul}

\floatname{algorithm}{Algorithm}

\title{Multi-Depth Uniform Coverage Path Planning for Unmanned Surface Vehicle Surveying}

\author{Maider Larrazabal$^{1,3}$~\IEEEmembership{Student Member,~IEEE}, Tong Yang$^2$~\IEEEmembership{Member,~IEEE}, \\
Izaro Goienetxea$^3$~\IEEEmembership{Member,~IEEE}, Jaime Valls Miro$^{1,4}$~\IEEEmembership{Member,~IEEE} \\
\thanks{$^1$ AZTI Foundation, Bizkaia, Spain. This work is contribution nº 1298 from AZTI, Marine Research, Basque Research and Technology Alliance (BRTA). Maider Larrazabal´s doctoral program is sponsored by the IKERTALENT programme, Dept. of Economic Development, Sustainability, and Environment, Basque Govt. Jaime Valls Miro is also with IKERBASQUE, Basque Foundation for Science, Bilbao, Spain. 
}
\thanks{$^2$ Dept. of Comp. Sc. and Technology, Tsinghua University, P.R. China.}
\thanks{$^3$ Dept. of Comp. Sc. and Artificial Intelligence, University of the Basque Country (UPV/EHU), Donostia-San Sebastián, Spain.}
\thanks{$^4$ Robotics Institute, University of Technology Sydney, NSW, Australia}
\thanks{ \textit{(Corresponding author:} Maider Larrazabal 
{\tt\footnotesize mlarrazabal@azti.es})}
}

\begin{document}

\maketitle
\thispagestyle{empty}
\pagestyle{empty}

\begin{abstract} 
This paper introduces a novel automatic coverage path planning algorithm for bathymetry surveying with unmanned surface vehicles. 
The detection range of the mapping sensor employed - a multibeam echo sounder - is heavily influenced by local seafloor depths. Hence, a path designed to uniformly cover the sea surface does not guarantee uniform coverage of the seafloor. Yet this is currently the typical process for bathymetric surveys, with the simplistic boustrophedon scheme along manually selected waypoints at constant depths being the most widespread planner used. 
The proposed scheme incorporates coarse prior depth information to pre-process the target region and adaptively guide path generation and sensing range configuration. 
By explicitly accounting for depth variations, the proposed algorithm designs a coverage path with optimised spacing between survey passes that adjusts the sensing beam aperture 
to achieve more consistent seafloor coverage. The proposed method is shown to offer significant improvements 
in both synthetic and real-world scenarios. Validations in challenging synthetic terrains achieves coverage ratios beyond 99\%, a marked improvement when compared with traditional boustrophedon paths revealing a maximum 75\% coverage. 
The same trend appears in realistic simulations using real bathymetric data from a coastal harbour, with coverage 
reaching over 92\%, and 
significantly surpassing boustrophedon sweeps with coverage rates below 65\%. 
Beyond improved performance, the scheme also brings a fully automated design, suitable for autonomous marine vehicles, thus offering practical utilities for real-world applications.    
\end{abstract}

\section{Introduction}
\label{sec:intro}

Bathymetric surveys of coastal areas entail mapping the underwater topography to obtain essential data for safe navigation, environmental monitoring, and habitat assessment. Traditionally, these surveys are conducted using manned vessels equipped with single or multibeam echo sounders (MBES). However, unmanned surface vessels (USVs) have emerged in recent years as a flexible and cost-effective alternative~\cite{kum2020application}~\cite{castano2024evolution}. 
USVs offer better accessibility to hard-to-reach areas, reduced operational costs, and enhanced safety by eliminating the need for onboard personnel. These advantages make USVs particularly attractive for high-resolution bathymetric surveys in dynamic coastal environments. 

\begin{figure}[t]
    \centering
    \begin{subfigure}[b]{0.42\columnwidth}
        \includegraphics[width=\columnwidth]{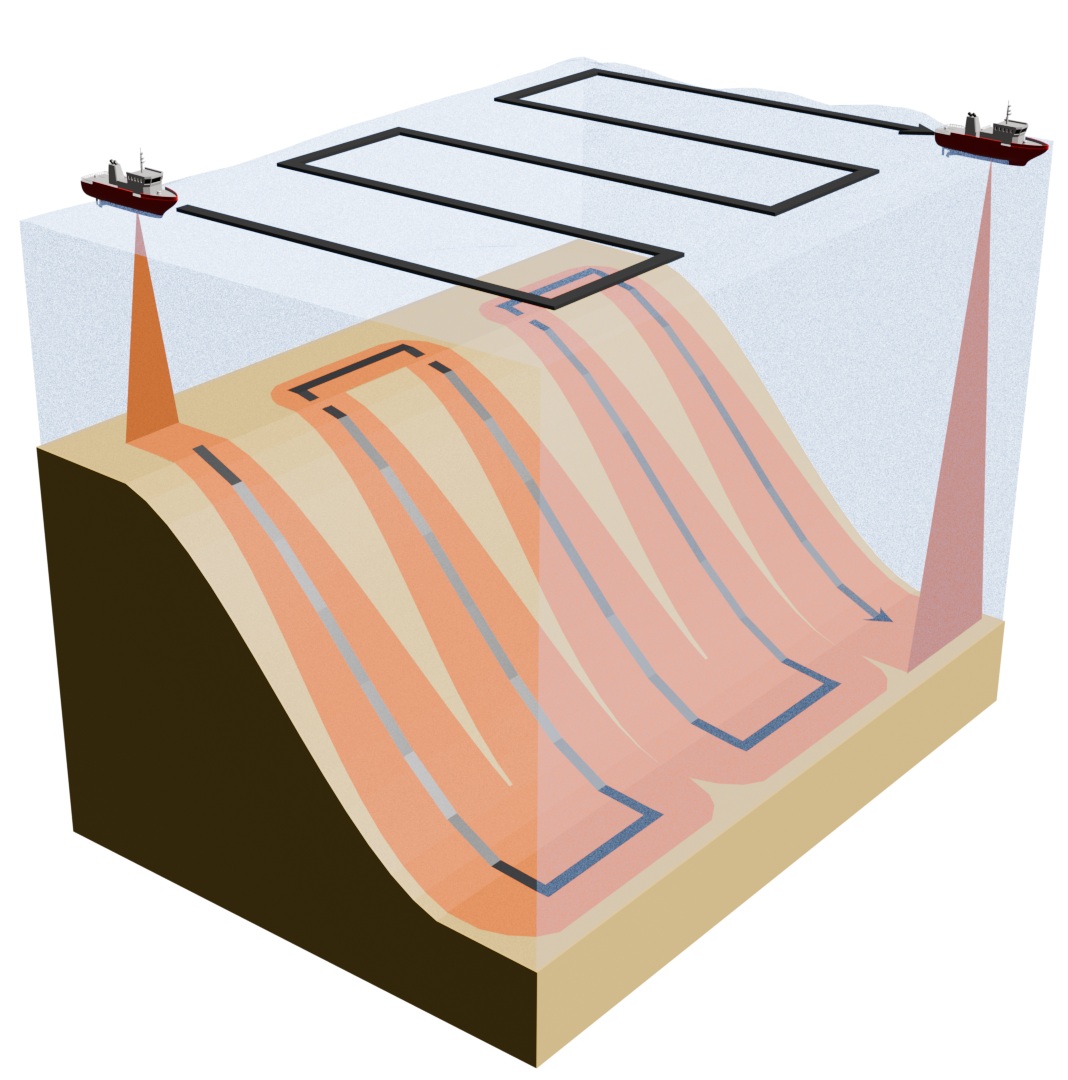}
        \caption{B\&F}
        \label{fig:baf concept}
    \end{subfigure}
    \quad    
     \begin{subfigure}[b]{0.42\columnwidth}
        \includegraphics[width=\columnwidth]{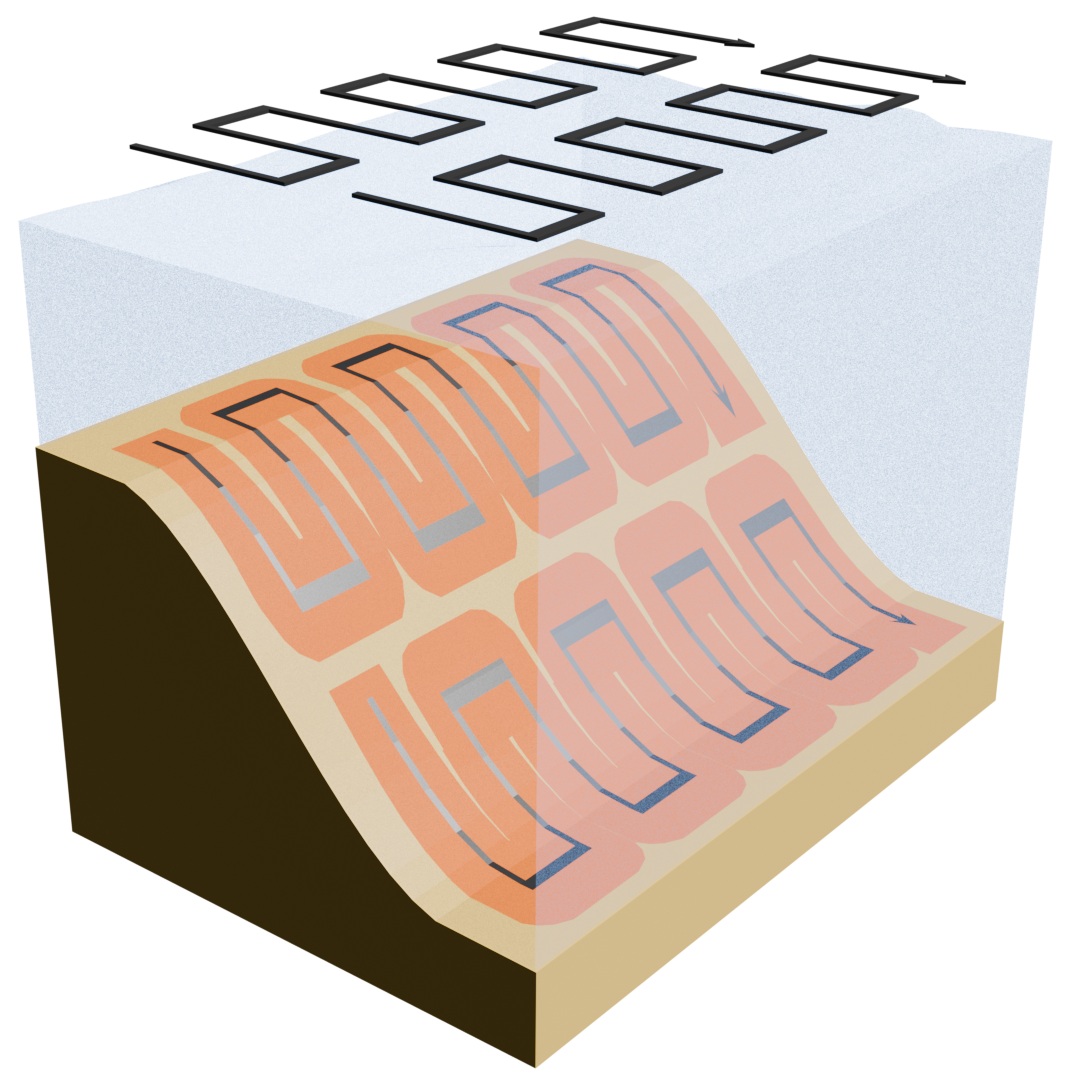}
        \caption{MDB\&F}
        \label{fig:mdbaf concept}
    \end{subfigure}
    \begin{subfigure}[b]{0.42\columnwidth}
        \includegraphics[width=\columnwidth]{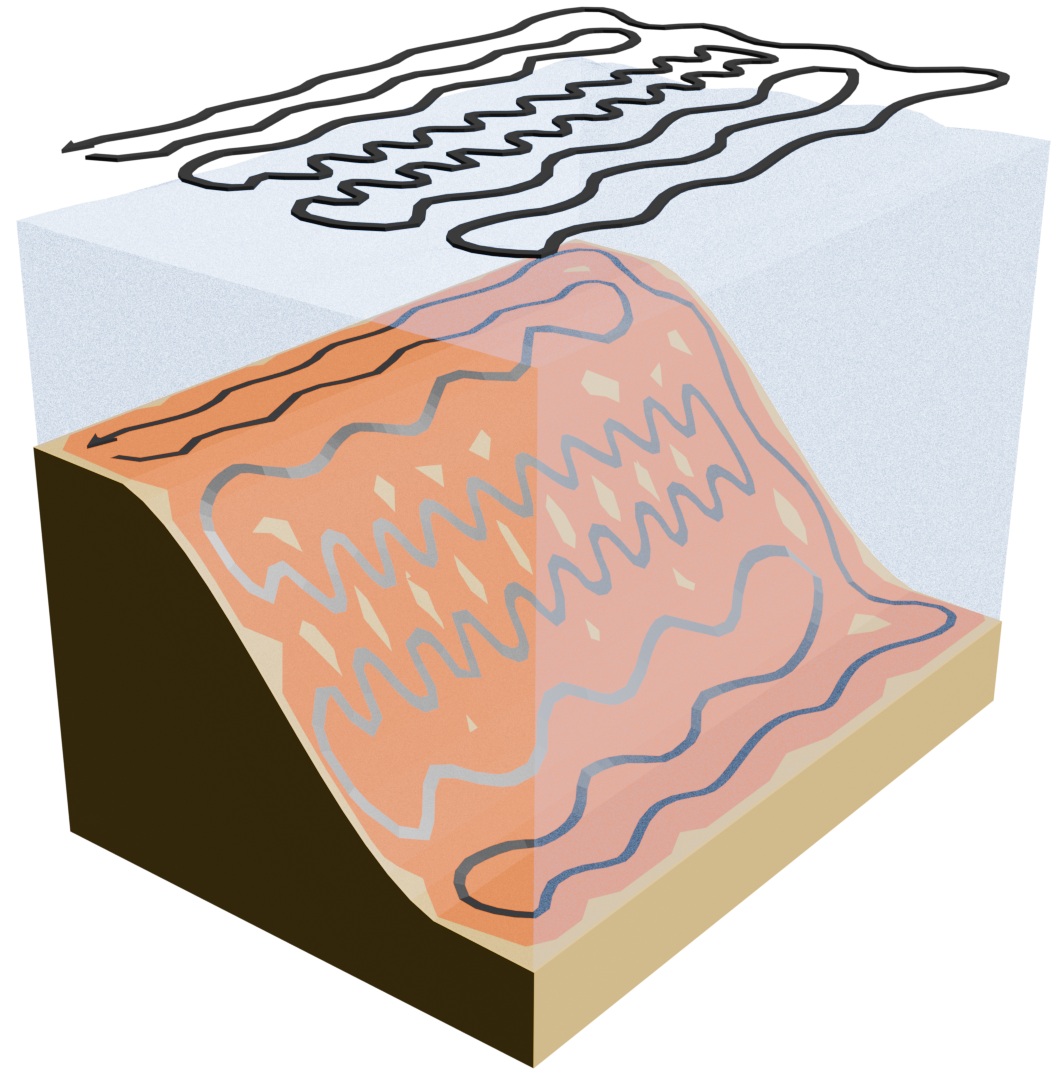} 
        \caption{MDNUC (Ours)}
        \label{fig:mdnuc concept}
    \end{subfigure}
    \caption{Coverage path planning for bathymetric surveys. a) Traditional Back-and-forth (B\&F) method, with constant sensor opening angle, generating gaps in the resulting coverage. b) Multi-depth B\&F (MDB\&F)), where area is separated into subregions according to seafloor depths (illustrated with 2 ranges), generating several paths with more constant footprint width and reducing coverage gaps. c) Multi-depth Non-revisiting Uniform Coverage (MDNUC), producing a single continuous path without resorting to cell subdivision, achieving highest coverage by optimally accounting for sensor footprint at varying survey depth ranges.}
    \label{fig:concepts}
    \vspace{-0.6cm}
\end{figure}

A fundamental component of any bathymetric survey is the coverage path planning (CPP)~\cite{galceran2013survey} algorithm, which determines how the vessel will navigate the region of interest (ROI) to collect complete and consistent data. Designing effective CPP strategies is essential to achieving key objectives such as optimizing study time, reducing energy consumption, and maintaining adequate data quality throughout the mission. This is a complicated task because previous studies have shown that CPP reduces to the well-known travelling salesman problem (TSP), an NP-hard problem in general. In addition to this important computational drawback, 
the CPP problem is further compounded by additional challenges in a maritime setting that make solving the problem even more challenging. A critical factor 
is the presence of static obstacles, such as islands, moored boats or docks, and dynamic obstacles, such as other vessels, which must be avoided. Additionally, the dynamic nature of the maritime environment introduces unique challenges, since USV like other watercrafts are subject to currents, waves and wind, causing it to more easily deviate from planned trajectory than ground vehicles. However, perhaps the most critical aspect of the CPP problem is the need for a physically feasible path that simultaneously addresses data quality requirements. This is a challenge that traditional coverage path planning methods have struggled to overcome.

\begin{figure}[t!p]
    \centering
    \includegraphics[width=0.8\columnwidth]{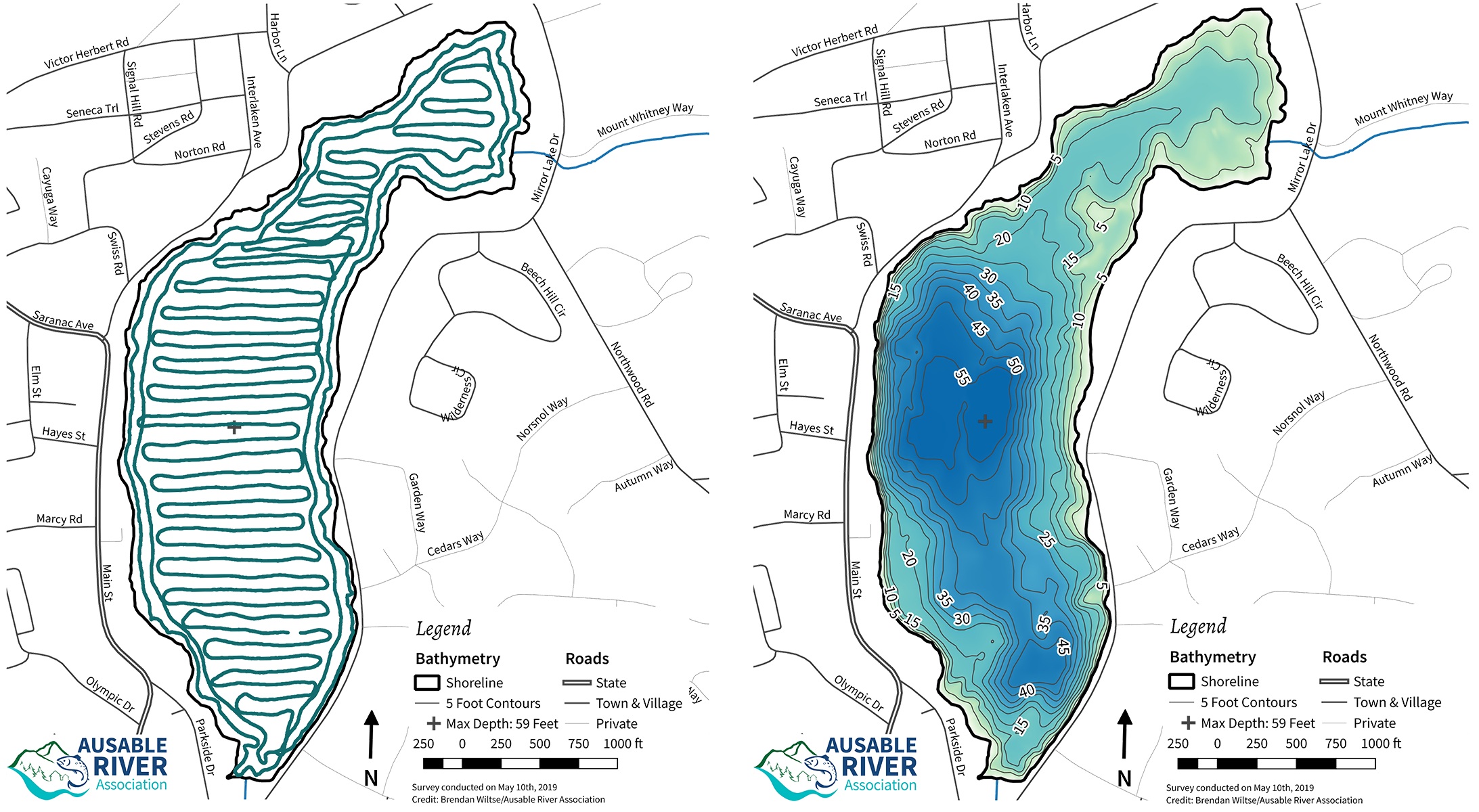}
    \caption{Bathymetric survey path and resulting depth map of Mirror Lake, NY, USA. Image from~\cite{ausable_bathymetry}.}
    \label{fig:mirror_lake_bathymetry}
    \vspace{-0.5cm}
\end{figure}

Traditionally, path planning for bathymetry has relied on methods based on geometric and fixed patterns, such as back-and-forth~\cite{Choset2000Coverage} (aka boustrophedon or lawnmower) or spirals~\cite{Cabreira2018Energy}. 
Whilst straightforward to implement, these approaches presume a constant sensor footprint, rendering them inefficient in real-world scenarios where water depths, thus sonar footprints fluctuate substantially. Fig.~\ref{fig:baf concept} shows the gap created when depth variation is not taken into account. This is a significant issue because, in practice, it reduces the applicability of these fixed patterns schemes to open, unobstructed areas with simple boundaries and relatively constant depths. For more complex topographies, the process often reverts to semi-manual supervision, where operators must adjust the sonar's beam width according to their experience with the sensor at known depths, reliance on other external measurements or drawing from local knowledge of the topography, manually guiding the USV to follow contours and ensure data overlap. A process that is both time-consuming and susceptible to human error. Fig.~\ref{fig:mirror_lake_bathymetry} shows a practical case where several passes had to be made manually, roughly following a boustrophedon style of parallel path lines to ensure complete coverage, making the process quite inefficient. As a result, there is considerable variability in the quality of coverage, often including substantial overlap and critical data gaps which end up being interpolated from the limited observed data, thus compromising the survey results.

Furthermore, many of these classical methods rely on cellular decomposition techniques (e.g., boustrophedon or morse decomposition)~\cite{galceran2013survey}~\cite{Choset2000Coverage}~\cite{acar2002morse} to divide the area into smaller subregions. While these techniques simplify route generation, they have significant limitations regarding computational cost and scalability. The process of defining these sub-cells is often arbitrary, and the transitions between them can be long and unoptimized, increasing mission time and energy consumption. The complexity of the decomposition process increases exponentially with environmental irregularity, which often results in significant planning time for large areas~\cite{Zhao2024Joint}). This sensitivity to parameter discretisation can limit the effectiveness of these methods for real-time or automated planning, and their adaptability to large geographic regions.

A straightforward cell decomposition approach that address the challenge posed by assuming a constant sensor footprint in areas with substantial depth variations consists of subdividing the region of interest (ROI) into different subregions based on varying depths. 
A back-and-forth path within each subregion can then be drawn to map the seafloor, selecting a different angle for the echo sounder footprint 
more in line with the perceived depth of the subregion seabed. This Multi-Depth Back-and-Forth (MDB\&F) strategy, illustrated in Fig.~\ref{fig:mdbaf concept}, indeed results in a reduction of gaps, clearly evident when compared with traditional back-and-forth methods relying on single-depth assumptions. 
However, as is often the case in most cellular decomposition schemes, 
the challenge of determining the coverage subregion sequence, and how to establish connections between routes to minimise overlap, is far from trivial. The selection of entry and exit points within each region significantly impacts the overall cost and efficiency of the route. 

This study presents a novel proposal for a CPP method for bathymetric surveys from a different perspective, based on the non-revisiting uniform coverage (NUC)~\cite{yang2023template} method. As MDB\&F, the proposed method is designed to adapt to the varying depth profiles present in the ROI during the planning process, dividing the region into subregions with different depths using information from previous bathymetric surveys. However, the calculated trajectory results in a single path that guarantees each subregion is visited only once. Hence, each time the USV enters a subregion, the opening angle of the MBES can be dynamically modified and the footprint width adjusted to the depth at each subregion, maintaining constant coverage throughout the full, single sweep. 
This approach offers a number of benefits over other methods. Firstly, it establishes a single path that covers the entire region, eliminating the need to optimize the order of visits to the different subregions. Secondly, it does not require cell decomposition, which improves scalability, simplifies implementation, and increases robustness in areas with irregular shapes or complex topography. As shown in Fig.~\ref{fig:mdnuc concept}, the use of a single path allows for comprehensive coverage, surpassing the effectiveness of the previously discussed methods.

The primary contributions of this work can thus be summarised as follows:
\begin{enumerate}
    \item A novel depth-aware CPP algorithm able to enhance coverage efficiency by adjusting the path based on seafloor depths, optimising sensor performance and minimising coverage gaps.
    \item A flexible and scalable method that eliminates the need for cellular decomposition, simplifying implementation and increasing robustness, thus making it highly suitable in complex or irregularly shaped survey areas.
    \item Performance validation through challenging synthetic datasets and real bathymetry data.
    \item The open-source release of the algorithm\footnote{Github: https://github.com/MaiLa24/mdnuc}.
\end{enumerate}

\section{Related Works}
\label{sec:related_works}

This section reviews related work in coverage path planning (CPP) and its applications to unmanned surface vehicle (USV) bathymetry surveying. 

\subsection{Coverage Path Planning}
The coverage path planning (CPP) problem has played a fundamental role in a wide range of robotics applications, including unmanned ground ~\cite{Badamasi2025Autonomous}~\cite{Yi2023Complete} and aerial vehicles~\cite{Cabreira2019Survey}~\cite{Jensen2020Near}~\cite{Bine2023Novel}, manipulators~\cite{Thakar2022Area}, and USVs~\cite{Xing2023Review}. 
The design of an effective coverage path is highly contingent upon the geometric and physical properties of the target surface~\cite{Luo2024Path}. 
Commonly adopted performance metrics for evaluating path quality include coverage rate~\cite{Zhao2024Optimal}, total path length~\cite{Sudha2024Coverage}, number of turns for back-and-forth patterns~\cite{Pan2025Optimal}, time to completion~\cite{Zhao2024Joint}, and energy consumption~\cite{Zhao2024Energy}. 

The CPP~\cite{galceran2013survey} problem was originally formulated for planar, multiply-connected regions, leading to the development of cellular decomposition approaches~\cite{Choset2000Coverage}~\cite{acar2002morse}. 
Subsequently, a large variety of alternative coverage path patterns were conceived~\cite{Lakshmanan2020Complete}, including spiral~\cite{Cabreira2018Energy}, dual-spiral~\cite{Wu2019Energy}, wavefront-based~\cite{zelinsky1993planning}, spanning-tree~\cite{gabriely2001spanning}, and fractal coverage~\cite{sadat2015fractal}. 
Later research efforts shifted towards developing algorithms for surfaces with increasingly complex geometries~\cite{Yang2020Cellular} to enable realistic robotic operations, such as coastline modelling~\cite{Zhao2024Optimal} and object surface polishing~\cite{Yang2020Cellular}. 
Target surfaces may be curved for instance, like a bridge pier requiring cleaning~\cite{Hassan2018Deformable}. 
Even simple-shaped surfaces may have coverage regions constrained by environmental factors revealing complex overall operations on a case-by-case basis~\cite{Glorieux2020Coverage}, as also seen in~\cite{Hassan2018Deformable} where coverage had to be carried out in arbitrarily shaped zones with marine growth. 
Robot accessibility can also be restricted by environmental obstacles in the workcell~\cite{Hassan2019Ppcpp} or terrain steepness~\cite{Wu2019Energy}, making naive geometry coverage path impractical for execution. 
Recently, it has been observed that cellular decomposition-based techniques usually face substantial challenges in finding appropriate decompositions for complex tasks, where homeomorphism-based mappings~\cite{Wu2019Energy} are introduced to fit a template coverage path onto arbitrary surfaces of equivalent connectedness. 
Alternatively,~\cite{yang2023template} proposed an algorithm that generates coverage paths in arbitrary shape surface based on a uniform surface representation, bypassing the need for cellular decomposition or homeomorphic transformation. 

\subsection{Bathymetric Seafloor Surveying}
The Unmanned Surface Vehicle (USV)~\cite{Hashali2024Route} coverage problem presents a unique hybrid challenge between 2D and 3D coverage. 
Whilst the vehicle operates on the 2D sea surface, the target seafloor surface is inherently 3D~\cite{Galceran2012Efficient} and multi-depth, often with only coarsely estimated bathymetry data, which itself motivates the need for detailed surveying~\cite{Sun2021Coastal}. 
Recent works in the realm of 2D USV coverage has seen an shift towards exploring collaborative multi-robot strategies to improve USV sea surface coverage for larger environments: For instance \cite{Deng2025Automatic} proposes a UAV-USV system for garbage collection, \cite{Jiang2025Collaborative} introduced a boustrophedon-contract network protocol (B-CNP) algorithm for multi-USV job partitioning in unknown environments, and~\cite{Ma2022Cciba} developed the CCIBA* algorithm to formulate collaborative behaviour strategies such as area division, area exchange, and obstacle recognition. 
Other contributions include~\cite{Tang2023Coverage} which combines walking-around movements with a biological inspired neural network (BINN) to help the USV reach obstacles' boundaries, and~\cite{Yang2023Cooperative}, which proposes a sequential algorithm for an array of USVs to survey multiple regions of interests.
In contrast, most recent advances in 3D coverage have evolved around their use in manipulators operating in free space~\cite{yang2023template}. These work are predicated on sensing footprints quite different to those employed in USV bathymetry, which tend to be non-circular, varying in size with depth and motion, thus complicating the assimilation of these state-of-the-art coverage approaches developed for disparate tasks such as contact surface manipulation. The result is standard 2D coverage patterns like back-and-forth or spiral motions performing suboptimally in depth-varying terrains, whilst full 3D coverage algorithms are inapplicable as USVs are constrained to operate in the 2D plane. 

This gap motivates the interest of this work in developing a practical 2D coverage framework tailored for USVs seeking effective seafloor coverage.

\section{Methodology}
\label{sec:methodology}
The proposed method is inspired by the NUC algorithm~\cite{yang2023template}, 
given its notable capacity to generate single, continuous coverage paths without the necessity to resort to arbitrary cellular decompositions, as well as its adaptability to all types of terrain, irrespective of their geometric complexity. These trademarks makes the planning solution suitably adept to be implemented on autonomous vehicles.

\subsection{NUC Planner}
\label{sec:nuc}
\begin{figure}
    \centering
    \includegraphics[width=\columnwidth]{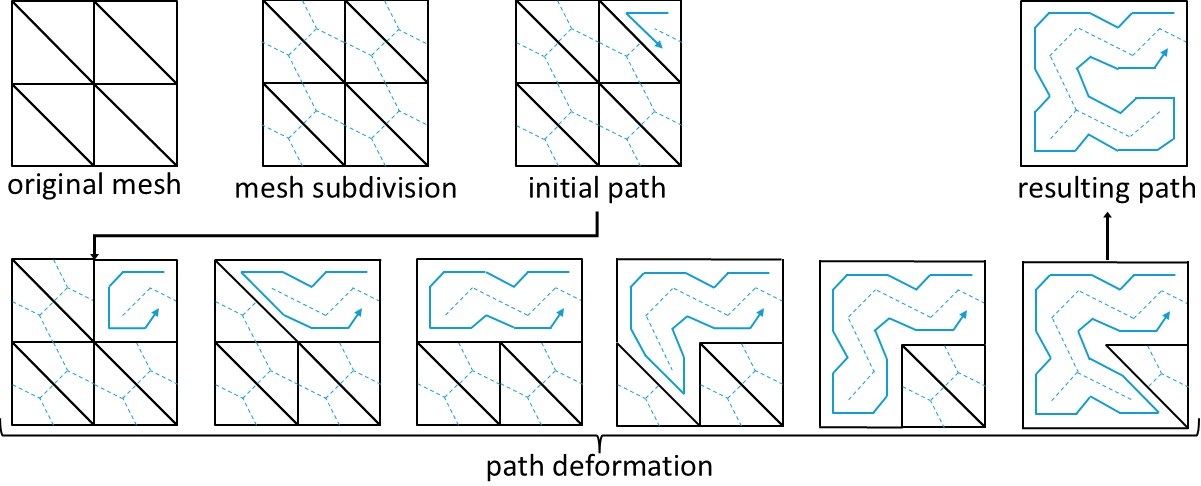}
    \caption{NUC path generation by face subdivision process.}
    \label{fig:nuc_path_generation}
    \vspace{-0.5cm}
\end{figure}

The NUC method was developed for a specific scenario that considers the area affected by a tool on a surface. The aim is to maximise the overall surface area covered while minimising the overlap between tool traces. The method firstly discretises the surface into an unstructured, uniform mesh. Then, a coverage skeleton is constructed directly from the unstructured mesh. This skeleton is a path that spans the surface without revisiting any point and is guaranteed to exist through mesh subdivision refinement, which divides facets to refine the mesh until the skeleton can be established. This ensures that path coverage always covers the entire mesh, regardless of its geometry complexity. A detailed, step-by-step illustration of the construction of the NUC resultant path is provided in Fig.~\ref{fig:nuc_path_generation}. For further details please refer to~\cite{yang2023template}. After generating the skeleton, local stochastic optimization can also be applied to improve coverage performance, reducing overlaps, gaps, and ensuring smoothness. This step fine-tunes the skeleton path to produce a more effective NUC path. Finally, for steady tracking, the path is generally interpolated into a smooth trajectory with constant speed, suitable for execution by the robot manipulator.

The NUC method has several advantages over traditional approaches. First and foremost, it does not use templates. This allows it to adapt to complex and irregular surfaces without relying on fixed geometric templates. Therefore, it can handle surfaces with holes and variable topologies without compromising coverage uniformity. Furthermore, the method ensures homogeneous coverage by refining the mesh and constructing skeletons, resulting in high-quality coverage. It can also be applied to unstructured meshes, making it suitable for complex real-world geometries. The process is fully automatic and robust, as it relies on local connectivity and mesh refinement rather than the aforementioned geometric templates, improving its applicability and reliability.

\subsection{Multi-Depth Non-revisiting Uniform Coverage (MDNUC)}
\label{sec:mdnuc}
While the NUC method provides a solid basis for uniform coverage, its original formulation is not directly applicable in other planning settings, such as bathymetric surveying. As mentioned, the planner is primarily intended for use by robotic arms with tools that are in constant contact with the surface to be covered. In the context of bathymetric surveys, the echo sounder ranging sensor, which is attached to the boat, and the surface to be inspected, which is located on the seabed, are clearly not in contact with each other. This work proposes a method to adapt the algorithm for use in the mapping of underwater seafloor surfaces, extending the NUC framework in three critical ways:

\subsubsection{Footprint-width Based Remeshing}
\label{sec:remeshing}
\begin{figure}
    \centering
    \includegraphics[width=0.35\columnwidth]{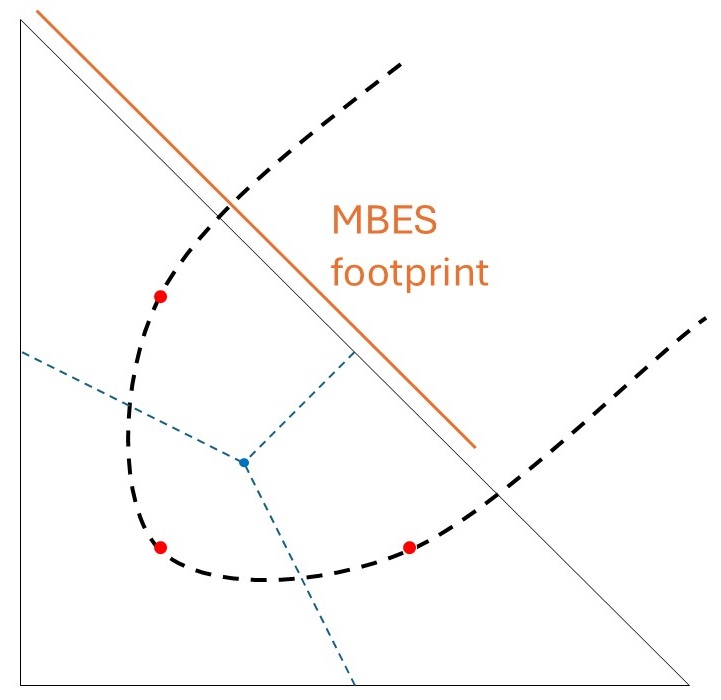}
    \caption{Mesh face geometry with example of resulting path.}
    \label{fig:footprint}
    \vspace{-0.6cm}
\end{figure}

In the original method, the mesh size is determined through experimentation, seeking a balance between not leaving areas uncovered and not covering more than necessary to avoid undue overlapping. 
NUC remeshes the surface of the initial mesh by dividing each triangle into three quadrilaterals using the centroid of the triangle as the separation point, as depicted in Fig.~\ref{fig:footprint}. The centroid of each quadrilateral  (depicted in red) is subsequently calculated, and the process repeats. When the subdivision process completes, the final centroid points are used to generate the coverage trajectory (shown as a dotted line in the figure).
For the bathymetric survey case, a different process must be followed driven by the physical characteristics of the MBES sensor footprint, and the desired bathymetry resolution (the spacing between depth measurements), with the aim to produce mostly isosceles right triangles during remeshing that can readily relate to these parameters. Specifically, the footprint width is calculated by multiplying the number of echo sounder beams by the selected resolution. This choice then drives the critical mesh size assumed for the remeshing process: since each edge is traversed exactly twice during planning, 
a length just over half the size of the face hypotenuse is equated to this desired sensor footprint, thus fully defining the remeshing parametrisation. The end result is a conservative choice intending to cover the whole face during the sensor sweep with minimal unnecessary data overlap. 

\subsubsection{Height-based Region Partitioning and Gating}
\label{sec:height_partition}
Since footprint width in the seafloor 
is directly tied to the distance of the sensor to the seabed, the mesh can be divided into height ranges. These ranges thus define regions within the mesh. The depth information gathered from the different subregions can then be used to automatically calculate the shared edges between them. Then, from all the shared edges between each pair of subregions, a single edge can then be selected as gateway between them, while the rest are defined as non-traversable.
Blocking all edges but one ensures continuity of the final coverage path along a single route 
between adjacent areas sitting at different surface depths.
In order to select the edge of the gate, the edge exhibiting the lowest slope is selected in order to minimize abrupt changes in depth. This ensures that the opening angle adapts more smoothly and efficiently when entering the new area to be inspected.

\subsubsection{Boundary Detection and Path Constraint}
\label{sec:boundary_detect}
After detecting the shared edges, the NUC algorithm receives information from both, non-traversable and gate edges. As highlighted above, the original algorithm traverses each mesh edge twice when generating a coverage path. However, the suggested MDNUC modification ensures that the path never intersects with blocked edges. Consequently, the only points of entry and exit between depth regions for the path are the gate edges, with a single gate edge designated for each pair of adjacent regions.
With this information, it is only necessary to modify the opening angle of the echo sounder once per region in order to adapt to the new region when crossing the gate.

\begin{figure}
    \centering
    \includegraphics[width=\columnwidth]{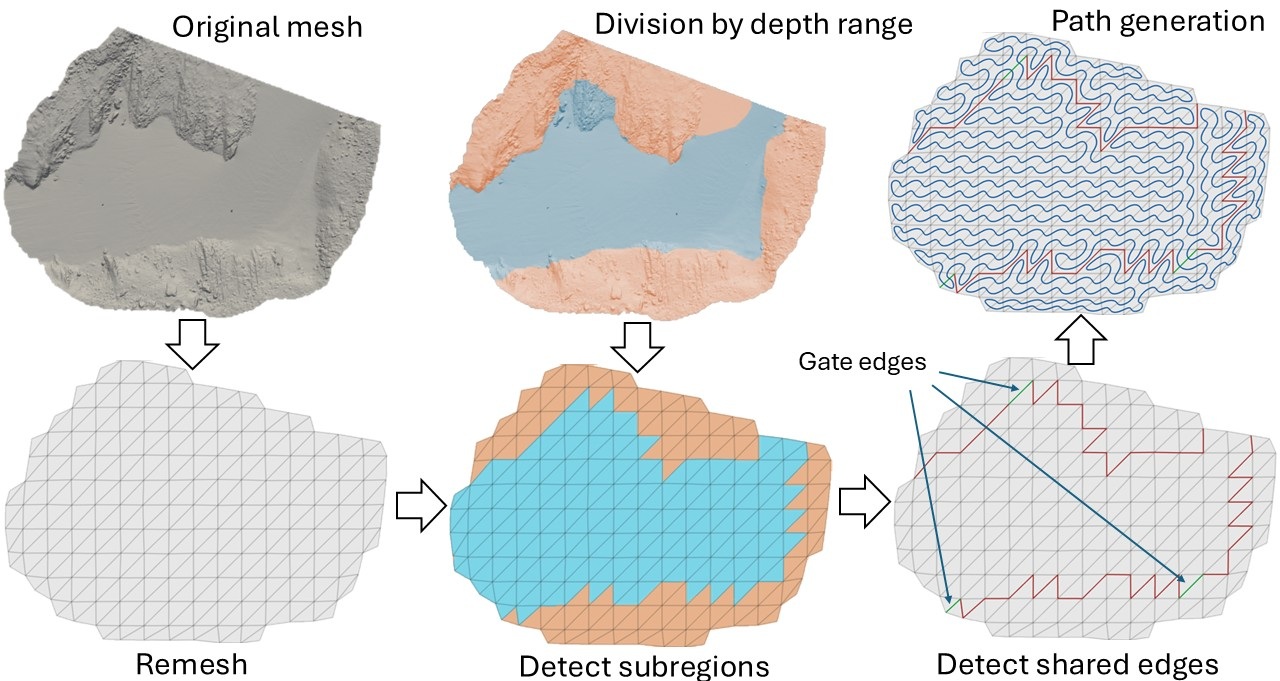}
    \caption{MDNUC Flowchart. 
    Example with two seabed depths ranges 
    (details in Table~\ref{tab:settings}), which result in four emerging subregions : three in shallower areas (orange hue; note the smaller one on the bottom left, delineated by a single face, with no common edges to the closeby orange section). And one in a deeper area (blue). Other regions would be revealed should depth ranges be chosen differently. Shared edges between subregions (red), with the three gates (green) that the path must intersect to access the different subregions. The final MDNUC path is depicted in blue, superimposed on the top right mesh, alongside the shared edges.}
    \label{fig:flowchart}
    \vspace{-0.5cm}
\end{figure}

Fig.~\ref{fig:flowchart} illustrates the overall MDNUC process, which involves the division of the survey area into subregions based on the depth of the seabed, the detection of blocked and gate edges, and the resulting path.

\section{Experiments}
\label{sec:experiments}
\begin{figure}
    \centering
        \begin{subfigure}[c]{0.45\columnwidth}
        \includegraphics[width=\textwidth]{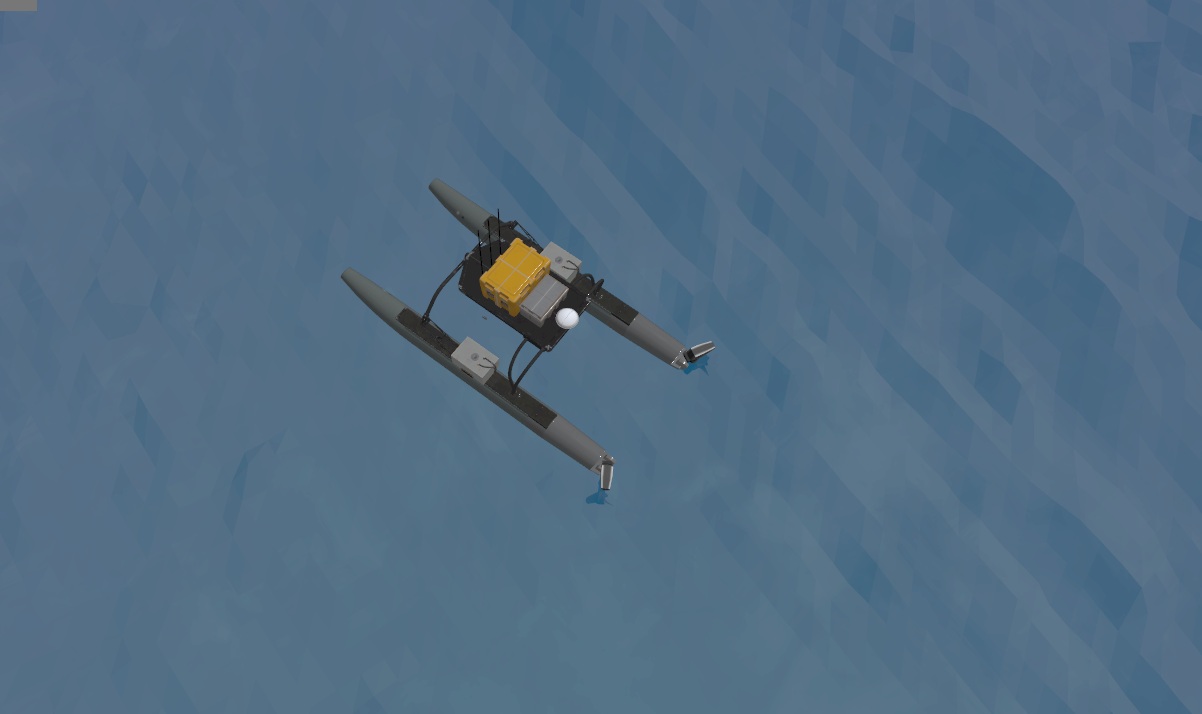}
    \end{subfigure}
    \begin{subfigure}[c]{0.53\columnwidth}
    \includegraphics[width=\textwidth]{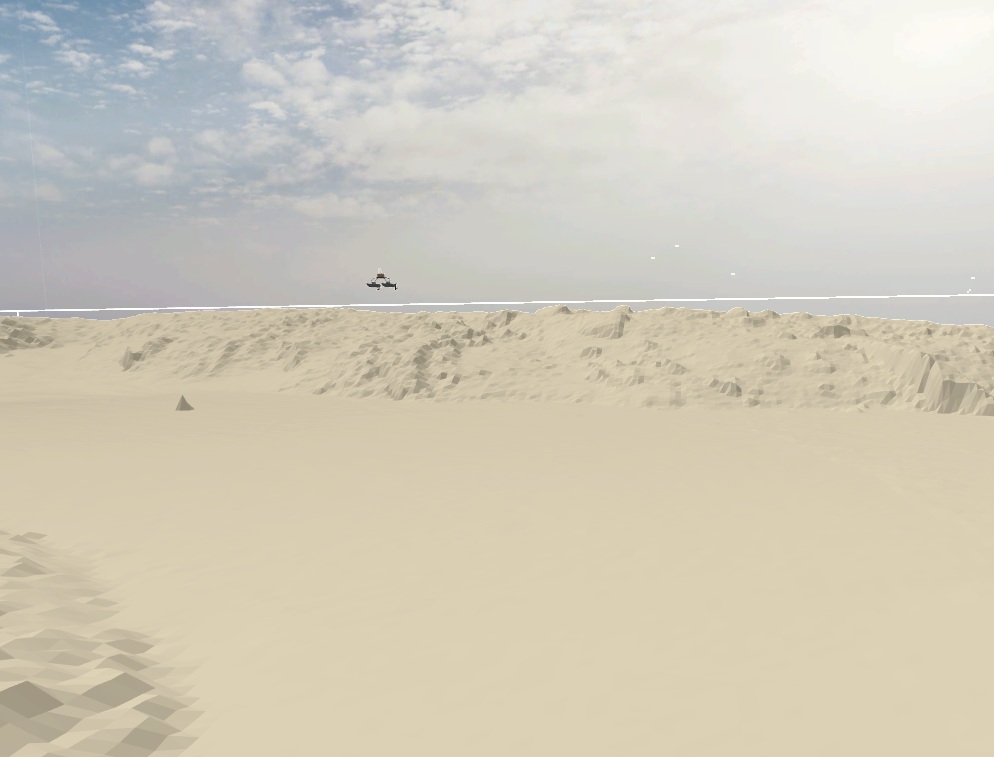}
    \end{subfigure}
    \caption{VRX ROS simulator, depicting the instrumented USV used in all the experiments and a detail of the Pasaia dataset.}
    \label{fig:usv}
    \vspace{-0.5cm}
\end{figure}

\begin{figure*}[t!p]
    \centering
     \begin{subfigure}[b]{0.22\textwidth}
        \includegraphics[width=\textwidth]{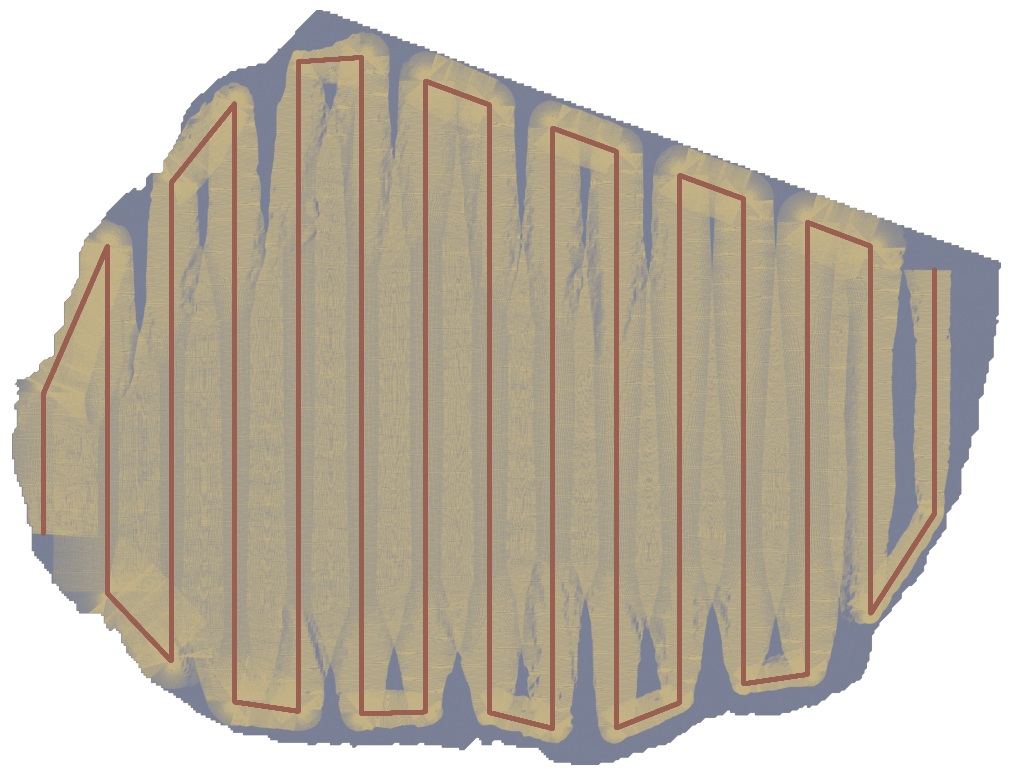}
        \caption{B\&F}
        \label{fig:pasaia_baf}
    \end{subfigure} 
    \quad
    \begin{subfigure}[b]{0.22\textwidth}
        \includegraphics[width=\textwidth]{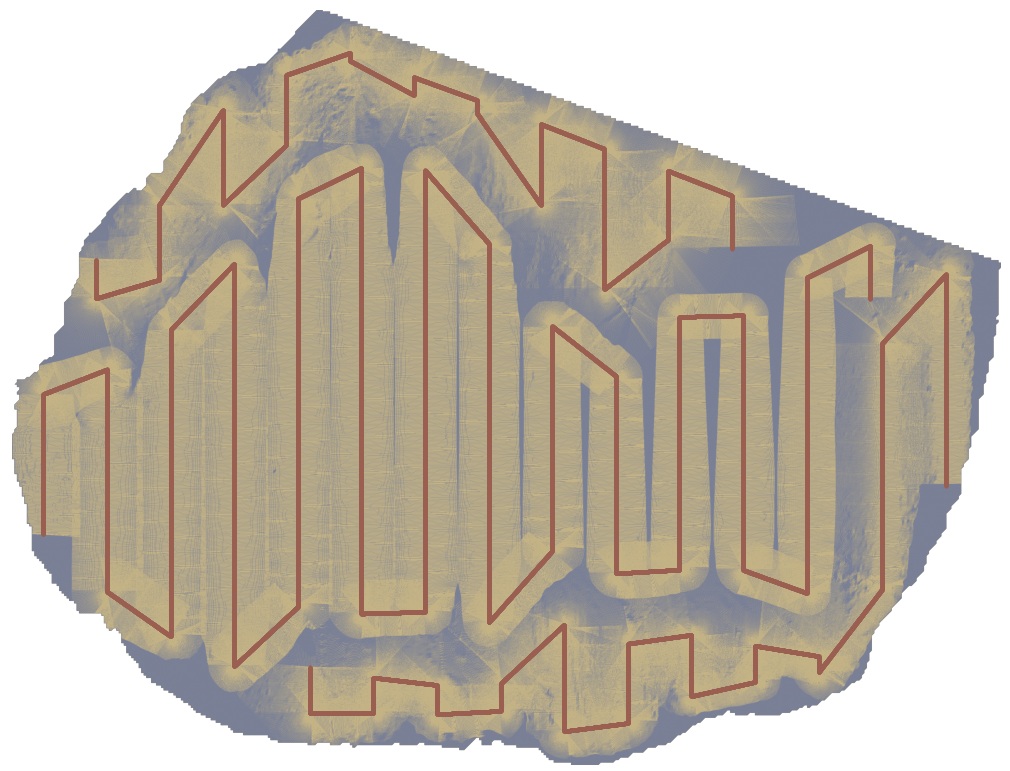}
        \caption{MDB\&F}
        \label{fig:pasaia_mdbaf}
    \end{subfigure}
    \quad
     \begin{subfigure}[b]{0.22\textwidth}
        \includegraphics[width=\textwidth]{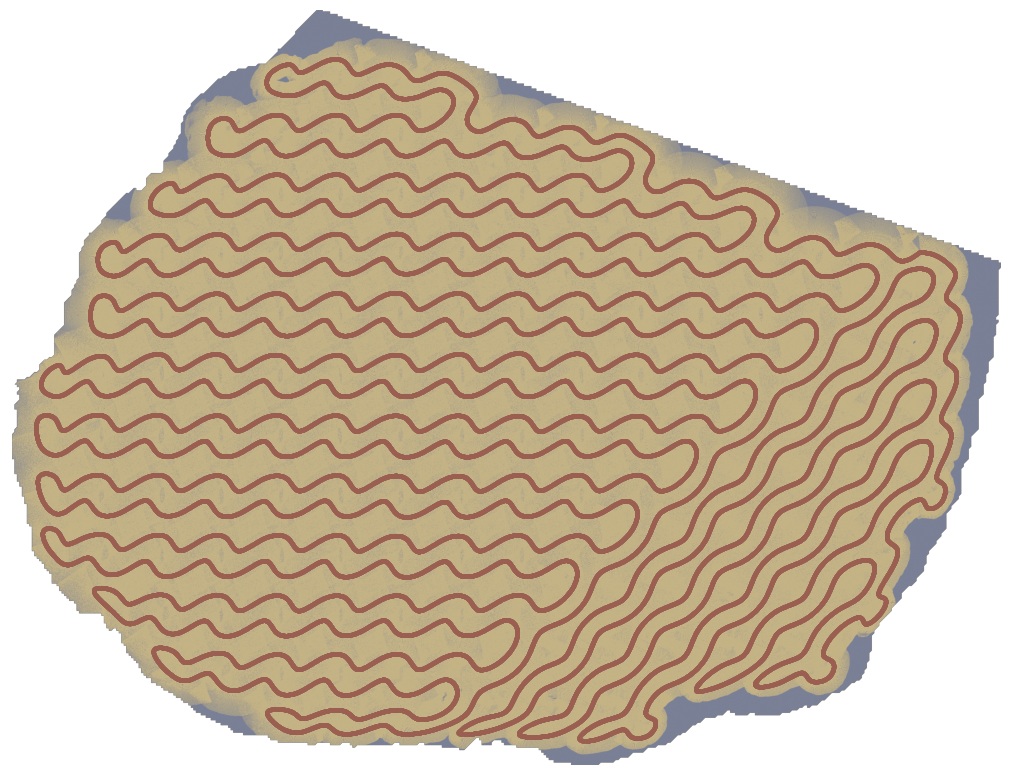}
        \caption{NUC}
        \label{fig:pasaia_nuc}
    \end{subfigure}
    \quad
    \begin{subfigure}[b]{0.22\textwidth}
        \includegraphics[width=\textwidth]{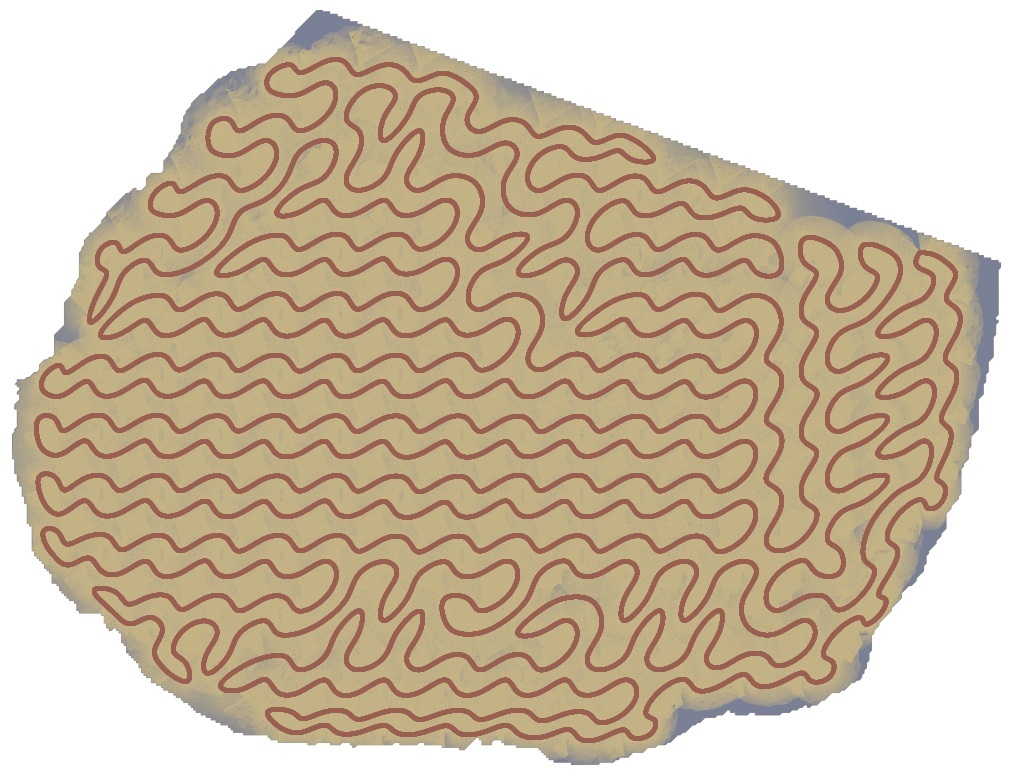}
        \caption{MDNUC}
        \label{fig:pasaia_mdnuc}
    \end{subfigure}    
    \quad
    \caption{Coverage mapping results on the Pasaia dataset (10 cm resolution).}\label{fig:results_concav}
    \vspace{-0.5cm}
\end{figure*}

\color{black}
To validate the performance and robustness of the proposed path planning method, a series of experiments were conducted, using both synthetic and real seafloor data collected from a bathymetric survey. The initial two experiments showcase the algorithm's capacity to manage intricate and challenging terrain structures, depicting both sharp and sloped terrains, while the third experiment substantiates its efficacy in dealing with data from a real-world scenario with a mix of seafloor topographies. All tests were performed in the ROS operating system environment by simulating the tracking of the resulting paths in the state-of-the-art VRX simulator for planning and control~\cite{bingham2019toward}, using the WAM-V USV model provided by the simulator (see Fig.~\ref{fig:usv} for an illustrative screenshot). The controller used during the experiments provided an average tracking error of 1.08m MAE and 1.18m RMSE in the largest tests carried out (the Pasaia real world scenario, following a 15 km path over an area of nearly 1 km$^2$, a section of which is shown in~\ref{fig:pasaia_mdnuc}), ensuring consistent execution of the planned trajectories.

The performance of MDNUC for seafloor mapping was compared with the original NUC method, the classic back-and-forth (B\&F) method, and the Multi-Depth B\&F (MDB\&F). The original NUC method is used to compare the coverage impact that the proposed MDNUC enhancements may bring forward against the perceived reference of a continuous path and cell subdivision-free planner, while the classic boustrophedon method is used as the state-of-the-art benchmark, being the standard practice in current bathymetric surveys. The application of MDB\&F (also developed as a simple comparative metric for this work) enables the assessment of MDNUC against another method with due consideration for depth range adaptability. However, because there is no straightforward mechanism to optimize the visiting order and where the angular change gates between the zones should be, the paths for each region have been created separately.

For the NUC and B\&F methods, the average height of the mesh was used to calculate the echo sounder's opening angle. For MDNUC and MDB\&F, the average heights of each mesh subregion were used. The width of the footprint remains constant in all methods, and the comparisons were carried out at two resolutions, 10 and 25 centimetres. Typically, bathymetric surveys utilise a resolution of 1 meter or above, with 25 centimetres reserved for enhanced detail. We have incorporated a finer 10-centimetre adjustment to further assess the precision of the methods. To evaluate coverage, a 2D grid with the desired resolution was generated, whereby a grid cell is considered covered if it is hit by a sonar beam.

\begin{figure}[t]
    \centering
    \begin{subfigure}[b]{0.32\linewidth}
        \includegraphics[width=\linewidth]{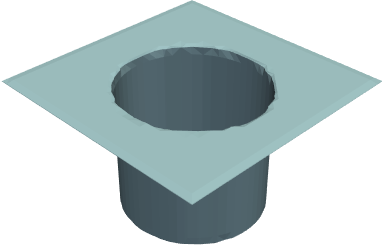}
        \subcaption{Shaft}
        \label{fig:concav_mesh}
    \end{subfigure}
    \begin{subfigure}[b]{0.32\linewidth}
        \includegraphics[width=\linewidth]{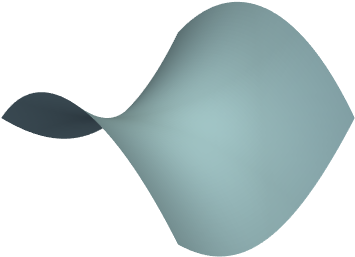}
        \subcaption{Saddle}
        \label{saddle_mesh}
    \end{subfigure}
    \begin{subfigure}[b]{0.32\linewidth}
        \includegraphics[width=\linewidth]{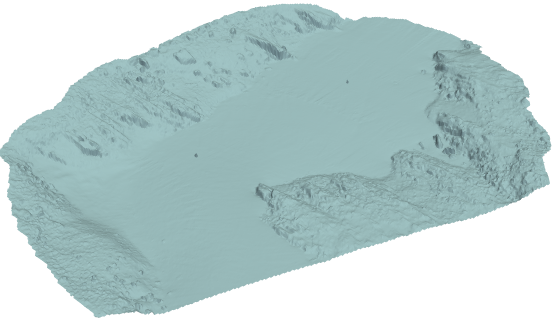}
        \subcaption{Real (Pasaia)}
        \label{pasaia_mesh}
    \end{subfigure}
    \caption{Synthetic and real meshes used for the experiments.}
    \label{fig:toy_examples}
\end{figure}

\begin{table}[t]
\centering
\caption{Planner coverage (\%) on synthetic scenarios.}
\label{tab:toy_results}
\begin{tabular}{lccccc}
\textbf{Seafloor}                & \textbf{Resolution} & \multicolumn{1}{l}{\textbf{B\&F}} & \textbf{MDB\&F} & \multicolumn{1}{c}{\textbf{NUC}} & \multicolumn{1}{c}{\textbf{MDNUC}} \\ 
\hline
\hline
\multirow{2}{*}{Shaft}   & 10 cm                &     30.63                              &   73.03              &  92.39                                &    \textbf{99.28}                              \\
& 25 cm                &          40.94                         &   66.93              &         92.93                         &      \textbf{98.10}                              \\ \hline
\multirow{2}{*}{Saddle} & 10 cm                &       67.08                            &    75.52             &    98.03                              &           \textbf{99.34}                       \\
& 25 cm                &           69.50                        &      63.42           &         93.64                         &     \textbf{94.90}                               \\ \hline
\end{tabular}
\vspace{-0.5cm}
\end{table}

\subsection{Synthetic Scenarios}
\label{sec:controlled_scenarios}
For the controlled examples with manufactured data, two meshes with challenging structures 
were created.
A concave shaft-like mesh (Fig.~\ref{fig:concav_mesh}) illustrates a scenario where there is a significantly deeper area in the centre of a plain. The objective is to evaluate the efficacy of the proposed method in scenarios involving sudden changes. For MDNUC and MDB\&F, two height ranges have been taken into consideration: one for the plain and another for the deeper area.

In contrast, the saddle surface (Fig.~\ref{saddle_mesh}) represents an area with significant yet smooth variations in height, which was used to assess the efficacy of changes in the echo sounder's opening angle. In this example, the mesh has been divided into four equal-height ranges.

The experimental results confirm the effectiveness of the proposed MDNUC method. As shown in Table ~\ref{tab:toy_results}, MDNUC consistently achieved the highest coverage in all scenarios. This is attributed to its ability by design to adapt route planning to the topology and depth of the terrain.

\begin{figure*}[t]
    \centering
     \begin{subfigure}[b]{0.22\textwidth}
        \includegraphics[width=\textwidth]{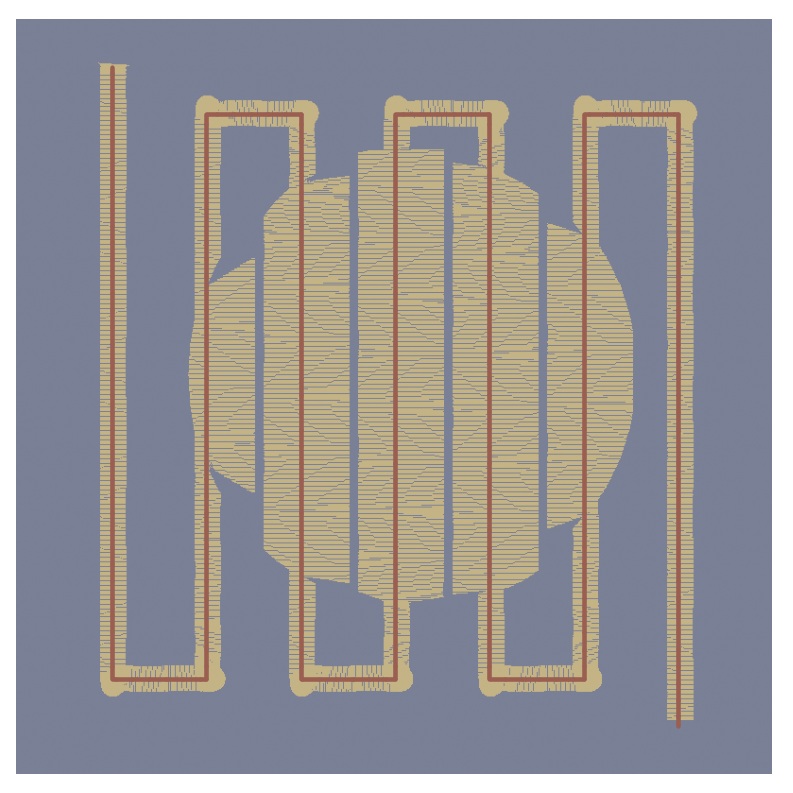}
        \caption{B\&F}
        \label{fig:concav_baf}
    \end{subfigure} 
    \quad
    \begin{subfigure}[b]{0.22\textwidth}
        \includegraphics[width=\textwidth]{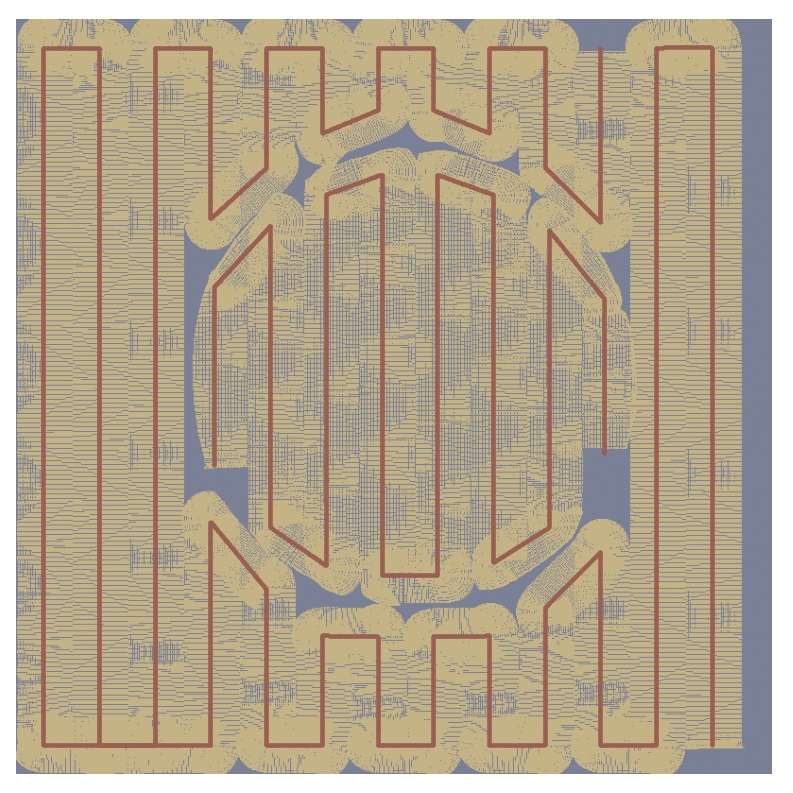}
        \caption{MDB\&F}
        \label{fig:concav_mdbaf}
    \end{subfigure}
    \quad
     \begin{subfigure}[b]{0.22\textwidth}
        \includegraphics[width=\textwidth]{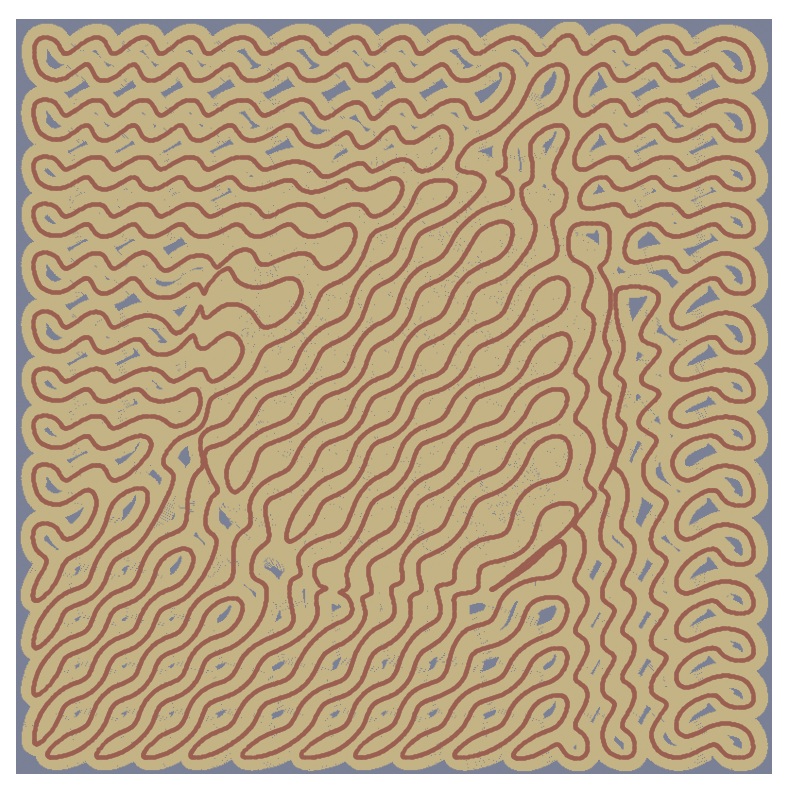}
        \caption{NUC}
        \label{fig:concav_nuc}
    \end{subfigure}
    \quad
    \begin{subfigure}[b]{0.22\textwidth}
        \includegraphics[width=\textwidth]{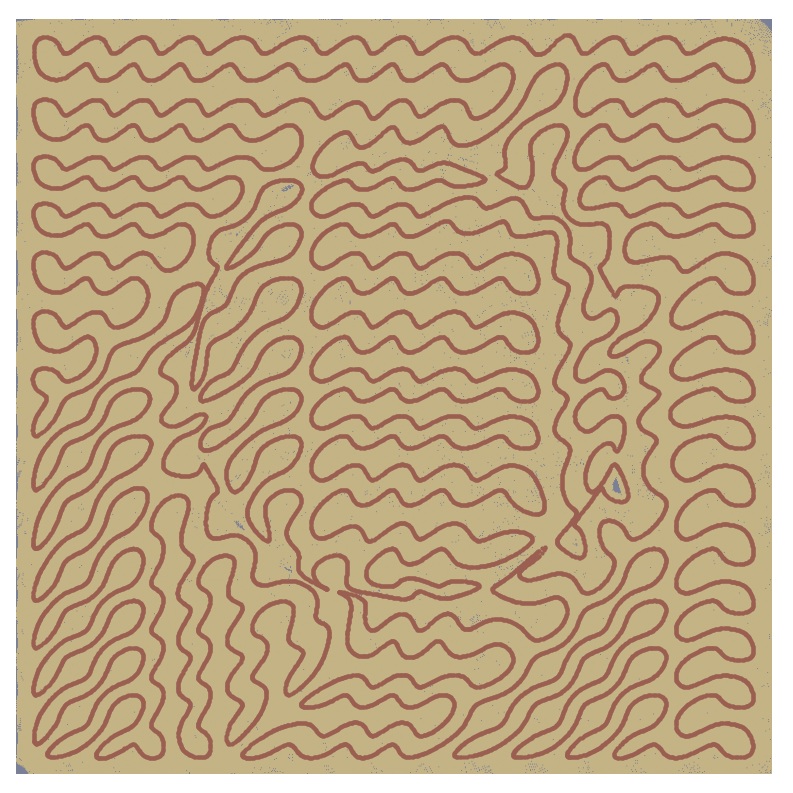}
        \caption{MDNUC}
        \label{fig:concav_mdnuc}
    \end{subfigure}
    \quad
    \caption{Coverage mapping results on the shaft dataset (10 cm resolution).}
    \label{fig:placeholder}
    \vspace{-0.3cm}
\end{figure*}

In the concave shaft scenario, the B\&F method (Fig.~\ref{fig:concav_baf}), with its constant opening angle, could only achieved limited coverage (30.63\% with a resolution of 10 cm). This outcome indicates its ineffectiveness in terrain with significant variations. Its fixed opening angle impedes the efficient sweeping of shallower areas, producing significant gaps as a result.

\begin{figure*}[t!p]
    \centering
     \begin{subfigure}[b]{0.22\textwidth}
        \includegraphics[width=\textwidth]{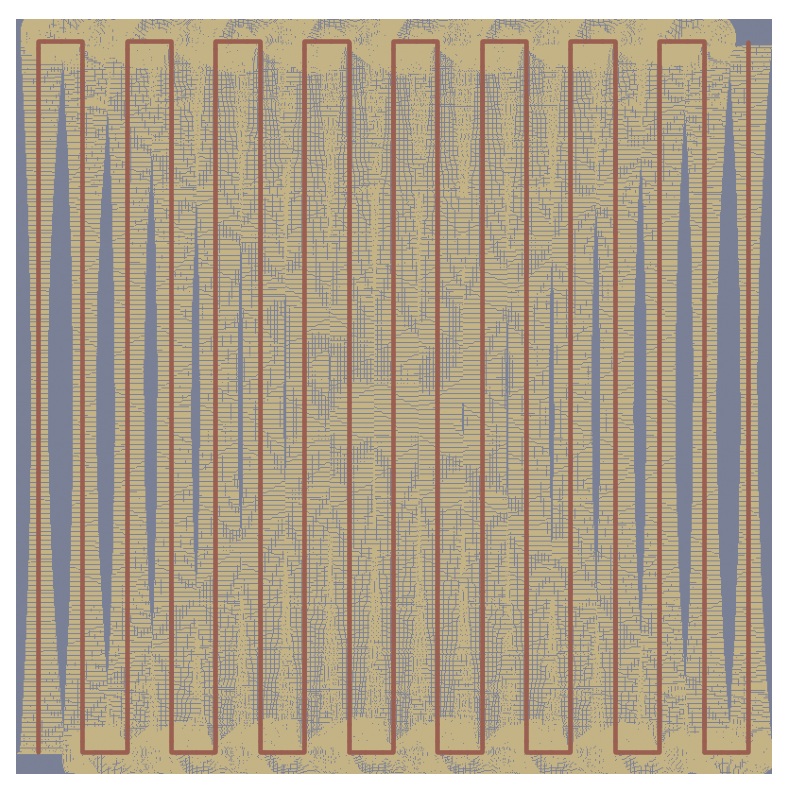}
        \caption{B\&F}
        \label{fig:saddle_baf}
    \end{subfigure} 
    \quad
    \begin{subfigure}[b]{0.22\textwidth}
        \includegraphics[width=\textwidth]{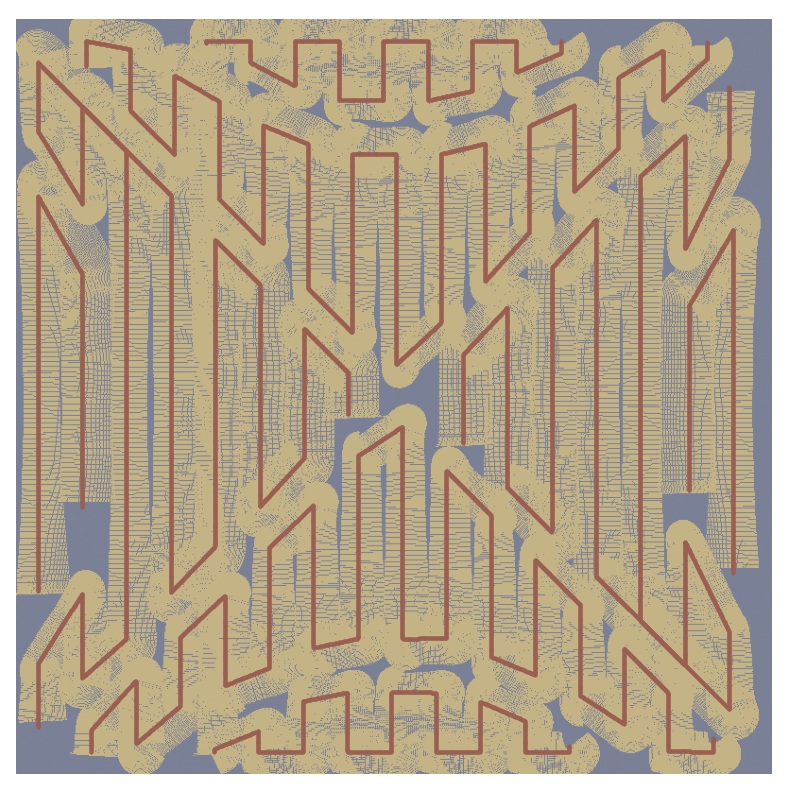}
        \caption{MDB\&F}
        \label{fig:saddle_mdbaf}
    \end{subfigure}
    \quad
     \begin{subfigure}[b]{0.22\textwidth}
        \includegraphics[width=\textwidth]{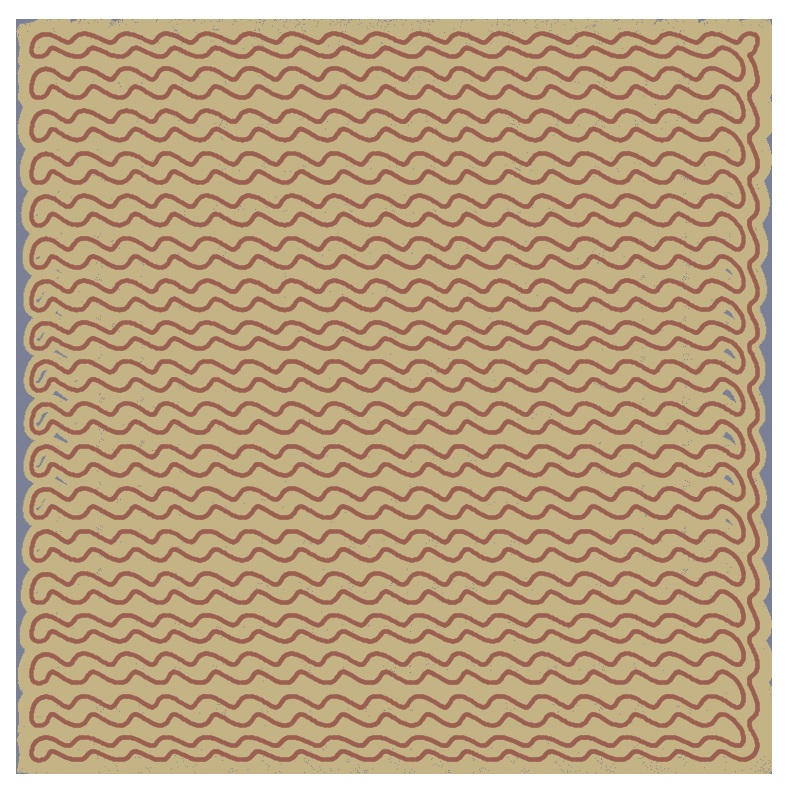}
        \caption{NUC}
        \label{fig:saddle_nuc}
    \end{subfigure}
    \quad
    \begin{subfigure}[b]{0.22\textwidth}
        \includegraphics[width=\textwidth]{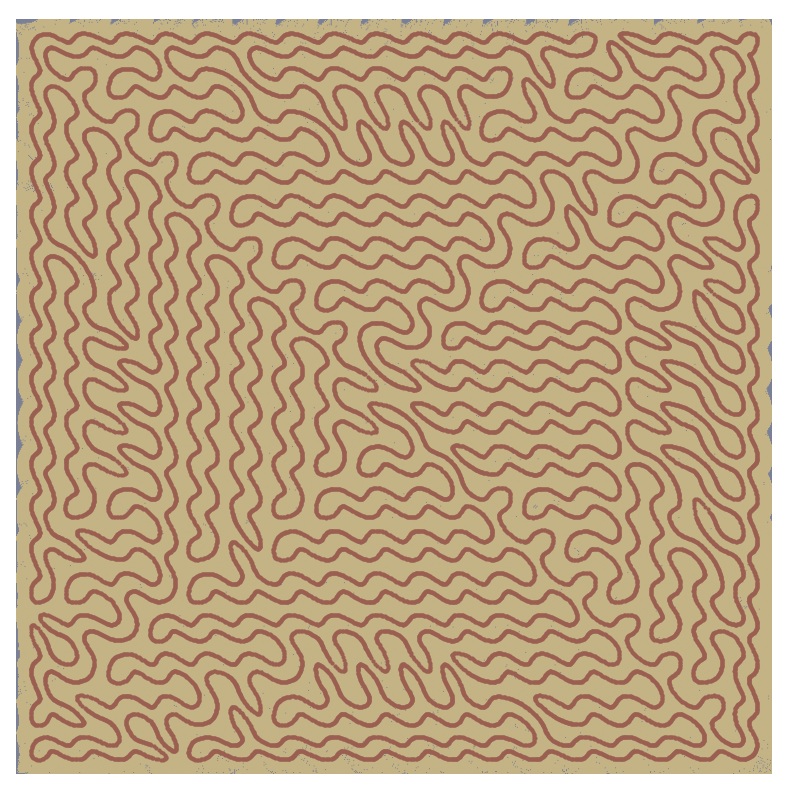}
        \caption{MDNUC}
        \label{fig:saddle_mdnuc}
    \end{subfigure}
    \quad
    \caption{Coverage mapping results on the saddle dataset (10 cm resolution).}
    \label{fig:results_saddle}
    \vspace{-0.5cm}
\end{figure*}

The MDB\&F method (Fig.~\ref{fig:concav_mdbaf}) uses a variable opening angle for each of the two depth zones, hence was able to achieve 73.03\% coverage at the same resolution. This result is considerably superior to that of B\&F, thereby validating the critical role that the height partitioning strategy plays in optimising coverage for bathymetry surveys. However, its linear scanning trajectory is not able to adapt to the seafloor surface topology, which effectively limits its ability to achieve complete coverage.

Notably, the NUC method (Fig.~\ref{fig:concav_nuc}) demonstrated 92.39\% coverage, surpassing the results achieved with B\&F and MDB\&F, despite maintaining a constant opening angle. This can be attributed to its ability to cover topologically complex areas. While this approach demonstrated enhanced robustness through its route planning, similar to the B\&F method, its fixed opening angle restricted its effectiveness in shallower areas, resulting in gaps that affected coverage.

Finally, the MDNUC method (Fig.~\ref{fig:concav_mdnuc}), which combines advanced, template-free NUC route planning with a variable opening angle for each zone, achieved an impressive 99.28\% coverage.  These results validate the central hypothesis of our research: combining template-free route planning and adapting the opening angle to depth is the most effective approach for bathymetry, demonstrating clear superiority over methods that use only one of these features.

On the saddle surface, NUC (Fig.~\ref{fig:saddle_nuc}) once again demonstrated its strengths by achieving 98.03\% coverage with a resolution of 10 cm, significantly outperforming B\&F  (Fig.~\ref{fig:saddle_baf}) with 67.08\%, and the MDB\&F method (Fig.~\ref{fig:saddle_mdbaf}) with 75.52\%. The performance of NUC in this scenario confirms its robustness as a route planning algorithm for irregular geometries.

The MDNUC method (Fig.~\ref{fig:saddle_mdnuc}) was still able to demonstrate slightly superior performance, achieving 99.34\% coverage. This marginal but noteworthy enhancement demonstrates the efficacy of the height partitioning and variable opening angle strategy. While NUC's route planning is already highly effective, the additional information from the different depths allows it to adjust the trajectory with greater precision, eliminating residual gaps and optimizing the route.

\subsection{Real-world Scenario}
\label{sec:real_scenarios}
\begin{figure}[t!p]
    \centering
    \includegraphics[width=0.69\columnwidth]{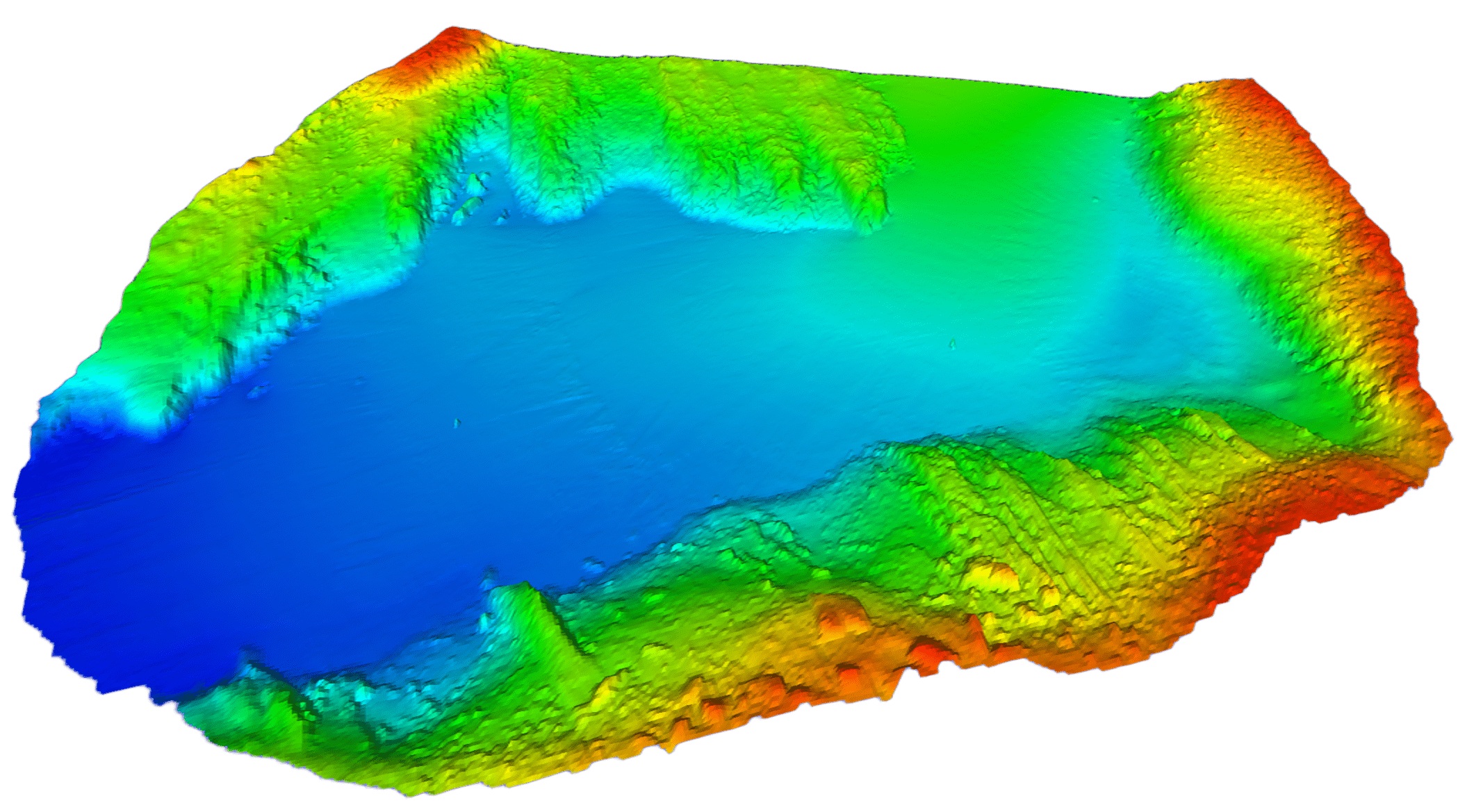}
    \caption{Higher resolution 3D bathymetry of Pasaia area (blue represents deeper seafloor sections, red shallower).}
    \label{fig:pasaia_detail}
    \vspace{-0.3cm}
\end{figure}

A 0.12 km² section of a seafloor survey (taken from Pasaia, a protected natural harbour in Gipuzkoa, Northern Spain), was selected for an experiment using real data from past bathymetric surveys (Figs.~\ref{fig:flowchart} and~\ref{pasaia_mesh}, also depicted in higher detail in Fig.~\ref{fig:pasaia_detail}). The area was selected to assess the effectiveness of the methods in a real and topologically complex environment. This section of the seafloor opens out to the sea through a channel with notable vertical wall constructs, hence representing a seafloor topology with sharper difference in height, ranging from 25.96 to 3.26 meters deep. 

The NORBIT iWBMS model (Norbit Subsea, Norway) was used as MBES reference. This system was selected for its ability to generate a high number of beams (configurable to 256 or 512) and for the flexibility of its configuration, which allows the beam opening angle to be modified during acquisition to optimize coverage at different depths. Moreover, this was also the sensing unit employed for the original data surveying.

In this example, the use of 256 beams has been chosen for both resolutions. For MDNUC and MDB\&F, two height ranges were defined: from 3.26 to 15 meters deep, and from 15 to 25.96 meters deep. For NUC and B\&F, the average was used as an acceptable compromise for a fair comparison, i.e., 14.6 meters deep. Table~\ref{tab:settings} collects the various values of the echo sounder's opening angle, considering the desired resolution and depth ranges.

\begin{table}[t!p]
\centering
\caption{MBES opening angle per depth range.}
\label{tab:settings}
\centering
\begin{tabular}{lcc}
\textbf{Depth Range (m)}               & \textbf{Resolution} & \textbf{Opening angle} \\ 
\hline
\hline
\multirow{2}{*}{3.26–15}               & 10 cm                & 122.65°                \\
    & 25 cm                & 155.32º                \\ \cline{2-3} 
    \multirow{2}{*}{15-25.96}              & 10 cm                & 60.38°                 \\
    & 25 cm                & 110.98º                \\ \cline{2-3} 
    \multirow{2}{*}{Mean depth $\sim14.6$} & 10 cm                & 82.44°                 \\
    & 25 cm                & 130.95º                \\ \hline
\end{tabular}
\vspace{-0.3cm}
\end{table}

As is evident in Table~\ref{tab:pasaia_results}, MDNUC consistently demonstrates its superior performance in a practical scenario also. In tests at 10 cm resolution, the B\&F (Fig.~\ref{fig:pasaia_baf}) and MDB\&F (Fig.~\ref{fig:pasaia_mdbaf}) methods achieved limited coverage - 64.81\% and 65.68\% respectively. This result underscores the inefficiency of a linear scanning trajectory for more complex areas. In contrast to the controlled scenarios, MDB\&F did not demonstrate notable improvement over B\&F in this scenario. A conclusion can be drawn that this is likely due to the added complexity of the real terrain, which necessitates continuous adaptation of the trajectory, not feasible under this scheme. The simple division into two depth zones is thus proven insufficient to achieve satisfactory results using this method. 

NUC (Fig.~\ref{fig:pasaia_nuc}), on the other hand, is able to demonstrate a substantial enhancement in its performance, achieving a coverage of 91.36\% and thereby confirming its suitability for complex geometries. However, its constant opening angle poses a limitation; with the inability of the strategy in dynamically adjusting to depth variations, gaps appear to intensify in shallower areas.

The MDNUC method (Fig.~\ref{fig:pasaia_mdnuc}) demonstrated superior performance, achieving 92.81\% coverage. The improvement, though seemingly minor, is noteworthy as it substantiates the efficacy of integrating NUC route planning with the division of the region into zones based on depth, improving over the NUC baseline that does not consider this. This approach enables the dynamic adjustment of the echo sounder's opening angle, a key advancement in our field.

The results at a resolution of 25 centimetres demonstrate a similar trend, and map depictions are withheld due to lack of space, with MDNUC achieving 81.97\% coverage in comparison to the lower results obtained by the other methods. In this case, the percentage improvement over the NUC baseline is indeed more significant. The reduction in coverage across all methods at this resolution was anticipated, as a coarser resolution tends to compromise the accuracy of route planning and coverage. However, the performance gap between MDNUC and the other methods widens, demonstrating that MDNUC is the most robust method even under conditions of lower mapping accuracy.

\begin{table}[t]
\centering
\caption{Planner coverage (\%) on real bathymetry data.}
\label{tab:pasaia_results}
\begin{tabular}{lcccc}
\textbf{Resolution} & \multicolumn{1}{l}{\textbf{B\&F}} & \textbf{MDB\&F} & \multicolumn{1}{c}{\textbf{NUC}} & \multicolumn{1}{c}{\textbf{MDNUC}} \\ 
\hline
\hline
10 cm                &        64.81                           &   65.68              &                        91.36           &                    \textbf{92.81}                  \\
25 cm                &        57.38                           &     60.53            &             74.71                     &                    \textbf{81.97  }              \\ \hline
\end{tabular}
\vspace{-0.5cm}
\end{table}

\section{Conclusion}
\label{sec:conclusions}

This paper presents the Multi-Depth Non-revisiting Uniform Coverage (MDNUC) algorithm, an innovative solution for USV bathymetric surveying. 
Conventional coverage algorithms - such as back-and-forth or spiral patterns - perform suboptimally in multi-depth marine environments, since a fixed sonar angle yields varying perceptual footprints on the seafloor depending on the underlying depth. 
To address this, the proposed algorithm leverages prior depth information to implicitly predict the effective sensing width of the sonar beams, and utilises the capability of modern sonar systems to adjust their angle of view on the fly accordingly. 
The target region is automatically partitioned based on local sea depth, effectively conforming an isocontour-based cellular decomposition without the need for any manual intervention. 
A template-free coverage path is then generated and executed in a cell-by-cell manner, ensuring efficient and complete coverage by a single sweeping path over the entire floor surface. These are key benefits for the advent of automated coverage schemes for robotic USVs. 
Experimental results show that the proposed algorithm achieves 99.28\% coverage in synthetic scenarios and 92.81\% coverage in the realistic Pasaia harbour area, a sharp performance improvement over standard current practices derived from boustrophedon-like algorithms.
An open source C++ and python implementation of the MDNUC software have also been released to support reproducibility and adoption by the community. 
There remains potential to improve path length and overlapping by refining the face remeshing process incorporating alternative shapes.

\bibliographystyle{IEEEtran}
\bibliography{references}

\end{document}